\documentclass[journal]{IEEEtran}
\ifCLASSINFOpdf
\else
\fi
\usepackage{amsmath}
\ifCLASSOPTIONcompsoc
	\usepackage[caption=false,font=normalsize,labelfont=sf,textfont=sf]{subfig}
\else
  \usepackage[caption=false,font=footnotesize]{subfig}
\fi
\usepackage{graphicx}

\usepackage{fixltx2e}

\usepackage{color}

\begin{document}
%
% paper title
% Titles are generally capitalized except for words such as a, an, and, as,
% at, but, by, for, in, nor, of, on, or, the, to and up, which are usually
% not capitalized unless they are the first or last word of the title.
% Linebreaks \\ can be used within to get better formatting as desired.
% Do not put math or special symbols in the title.
\title{Anchor Cascade for Efficient Face Detection}
%
%
% author names and IEEE memberships
% note positions of commas and nonbreaking spaces ( ~ ) LaTeX will not break
% a structure at a ~ so this keeps an author's name from being broken across
% two lines.
% use \thanks{} to gain access to the first footnote area
% a separate \thanks must be used for each paragraph as LaTeX2e's \thanks
% was not built to handle multiple paragraphs
%

\author{Baosheng~Yu~and~Dacheng~Tao,~\IEEEmembership{Fellow,~IEEE}% <-this % stops a space
\thanks{B.~Yu and D.~Tao are with the UBTECH Sydney AI Center, The University of Sydney, Darlington, NSW 2008, Australia, and also with the Faculty of Engineering and Information Technologies, School of Information Technologies, The University of Sydney, Darlington, NSW 2008, Australia (e-mail: bayu0826@uni.sydney.edu.au; dacheng.tao@sydney.edu.au).}% <-this % stops a space
\thanks{Manuscript received..}}

\maketitle

% As a general rule, do not put math, special symbols or citations
% in the abstract or keywords.
\begin{abstract}
Face detection is essential to facial analysis tasks such as facial reenactment and face recognition. Both cascade face detectors and anchor-based face detectors have translated shining demos into practice and received intensive attention from the community. However, cascade face detectors often suffer from a low detection accuracy, while anchor-based face detectors rely heavily on very large networks pre-trained on large scale image classification datasets such as ImageNet \cite{russakovsky2015imagenet}, which is not computationally efficient for both training and deployment. In this paper, we devise an efficient anchor-based cascade framework called anchor cascade. To improve the detection accuracy by exploring contextual information, we further propose a context pyramid maxout mechanism for anchor cascade. As a result, anchor cascade can train very efficient face detection models with a high detection accuracy. Specifically, comparing with a popular CNN-based cascade face detector MTCNN \cite{zhang2016joint}, our anchor cascade face detector greatly improves the detection accuracy, e.g., from 0.9435 to 0.9704 at 1k false positives on FDDB, while it still runs in comparable speed. Experimental results on two widely used face detection benchmarks, FDDB and WIDER FACE, demonstrate the effectiveness of the proposed framework.
\end{abstract}

% Note that keywords are not normally used for peerreview papers.

% For peer review papers, you can put extra information on the cover
% page as needed:
% \ifCLASSOPTIONpeerreview
% \begin{center} \bfseries EDICS Category: 3-BBND \end{center}
% \fi
%
% For peerreview papers, this IEEEtran command inserts a page break and
% creates the second title. It will be ignored for other modes.
\IEEEpeerreviewmaketitle

%%%%%%%%%%%% INTRODUCTION %%%%%%%%%%%%%%%%%%%%
\section{Introduction}

\IEEEPARstart{F}{ace} detection, aiming to locate all faces in a given image, is indispensable to a variety of facial analysis tasks, such as facial point detection \cite{Sun_2013_CVPR}, facial reenactment \cite{thies2015real,thies2016face2face}, and face recognition \cite{sun2014deep, schroff2015facenet,parkhi2015deep}. While the seminal cascade face detector \cite{viola2004robust} showed impressive performance for real-time near-frontal face detection, face detection in real-world scenarios remains very challenging due to large variations in scale, illumination, expression, pose, and occlusion \cite{yang2016wider}.

The cascade framework is suited to efficient face detection due to its fundamentals, i.e., an extremely unbalanced binary classification problem with a majority of negative samples \cite{viola2004robust}. A cascade face detector rejects most of easy negative samples in the early stages using simple classifiers. As a result, cascade face detectors are generally extremely fast \cite{wu2008fast}. To further boost cascade face detector performance in real-world scenarios, robust features are crucial for detecting non-frontal faces with large variations in scale and occlusion \cite{zhang2007face}. Recently, convolutional neural networks (CNNs) have been effectively applied to image classification \cite{krizhevsky2012imagenet} due to their great capacity for representation learning \cite{bengio2013representation}. Inspired by this, CNNs have been introduced to traditional cascade face detectors by replacing the original node classifiers, e.g., AdaBoost, with CNNs \cite{li2015convolutional}. However, cascade face detectors detect faces of different scales using a dense image pyramid and the computations on different slices of the dense image pyramid are not shared. As a result, CNN-based cascade face detectors use only tiny CNNs (e.g., two or three convolutional layers) in early stages, significantly degrading their performances in real-world scenarios \cite{li2015convolutional,qin2016joint,zhang2016joint}.

\begin{figure}[!t]
\centering
\includegraphics[width=0.5\textwidth]{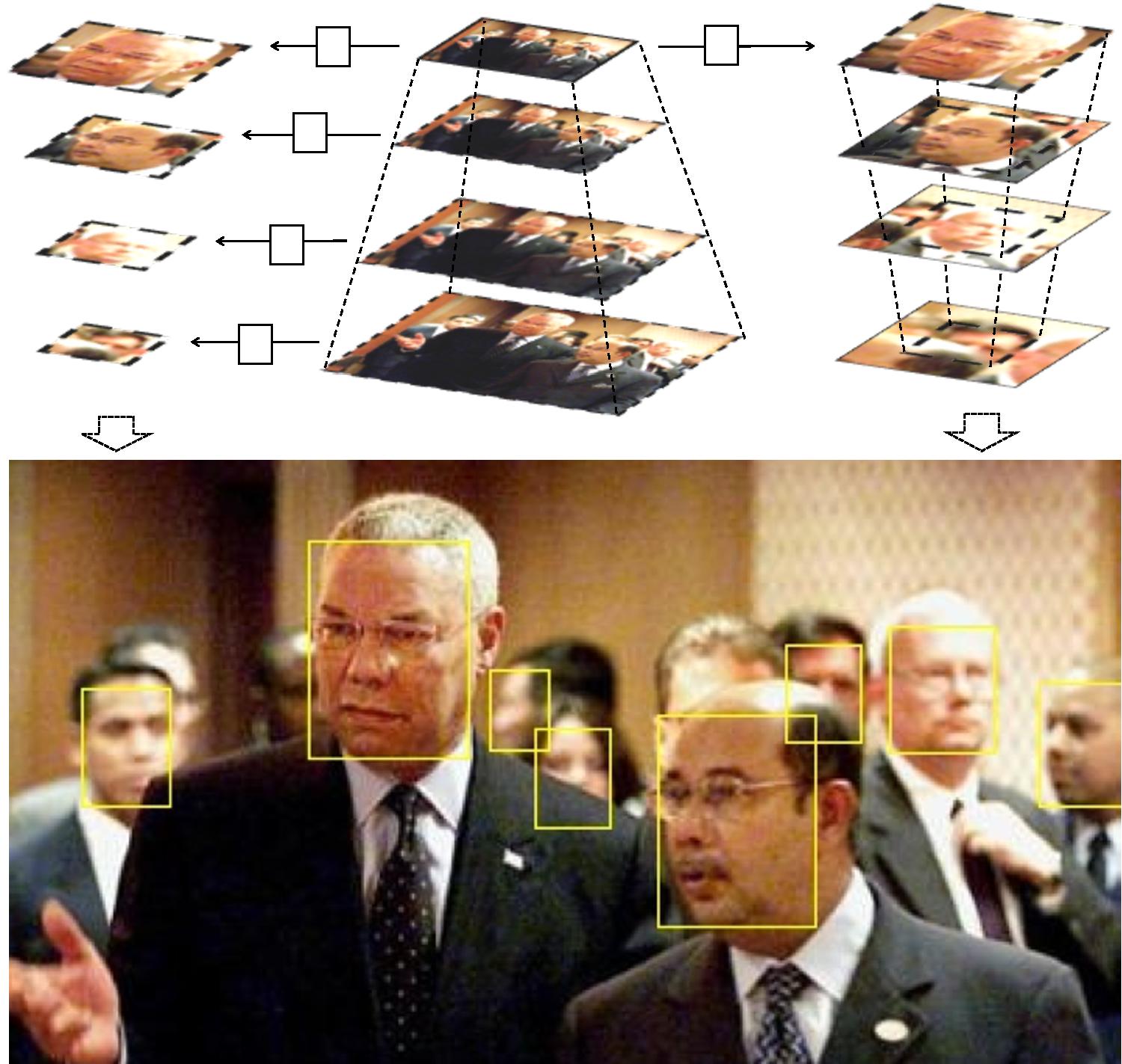}
\caption{A comparison between the traditional single-scale detector (left) and the anchor-based detector (right). To detect faces of different scales, a single-scale detector needs to scan an image pyramid slice by slice, while an anchor-based detector only needs to scan the smallest slice of the same image pyramid. By using the anchors, the computational cost are greatly reduced. For example, given the scale factor $\alpha=0.7937$ for the image pyramid and a set of $4$ anchors, the computational cost of anchor-based detector will reduce to approximate $1/10$ comparing with the single-scale detector. }
\label{fig:apn:demo}
\end{figure}

\begin{figure*}[!t]
\begin{center}
   \includegraphics[width=1.0\linewidth]{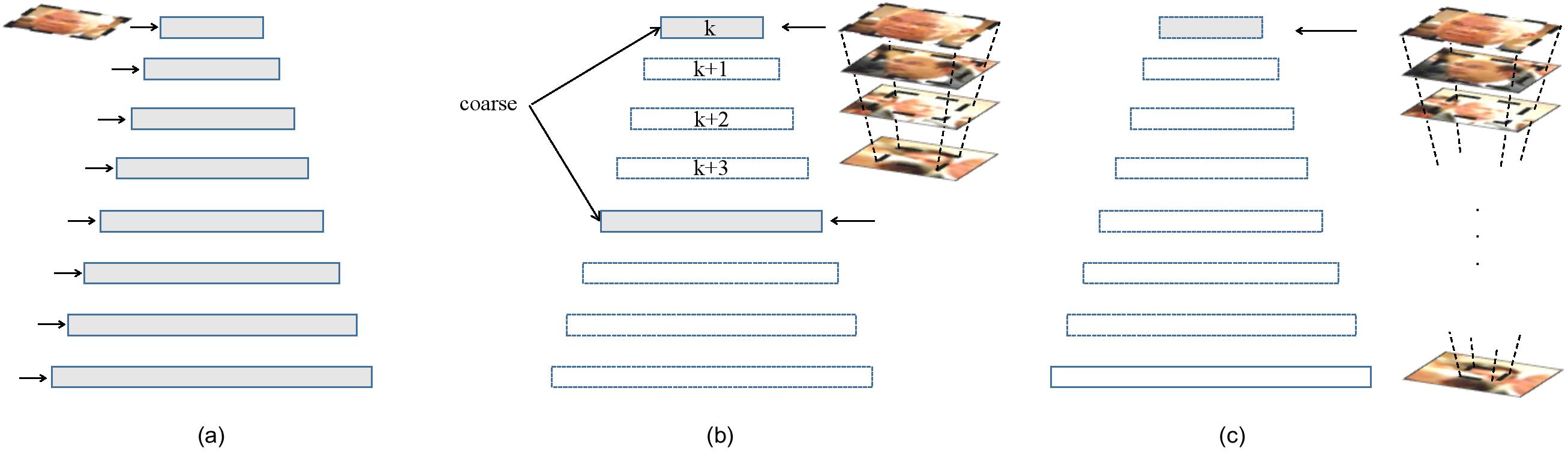}
\end{center}
   \caption{An example of the trade-off between the multi-scale anchors and the dense image pyramid for dealing with large variations in object scale. (a) A single-scale detector with a dense image pyramid, i.e., traditional cascade face detectors rely on a dense image pyramid to detect faces of different scales. (b) a set of anchors with a coarse image pyramid, which is a computationally efficient. (c) A full set of anchors without an image pyramid, i.e., anchor-based detectors rely on multi-scale anchors to detect faces of all scales.}
\label{fig:coarse:pyramid}
\end{figure*}

Anchor-based detectors such as Faster R-CNN \cite{ren2015faster} and single shot multibox detector (SSD) \cite{liu2016ssd} have become the most popular frameworks for generic object detection. With the help of multi-scale anchors, anchor-based detectors can detect objects of different scales simultaneously, and thus work without an image pyramid. We demonstrate a comparison between the traditional single-scale detector and the anchor-based detector in Fig.~\ref{fig:apn:demo}. Anchor-based detectors usually use anchors to cover an extremely wide range of object scales, e.g., from 16x16 pixels to 512x512 pixels. As a result, very large CNNs, e.g., VGGNet \cite{Simonyan14c} and ResNet \cite{he2016deep}, are usually indispensable for anchor-based detectors to learn scale-invariant features. Furthermore, large CNNs usually need to be pre-trained on a large-scale image classification dataset such as ImageNet \cite{russakovsky2015imagenet}, which takes intensive computational load and makes it difficult for us to explore new network architectures for detection. As a special case of generic object detection, face detection has been dominated by anchor-based detectors on detection accuracy \cite{jiang2017face,hu2016finding,najibi2017ssh,zhang2017s3fd}. However, anchor-based detectors rely heavily on very large base networks, which have become the bottleneck for efficient face detection in both training and inference phases.

For efficient face detection, it is important to avoid both a dense image pyramid and very large base networks. Two observations raise our concerns: (1) anchor-based detectors use very large CNNs to handle extremely large variations in object scale; and (2) the computational cost for scanning a dense image pyramid has a long-tail property, i.e., the majority of computational cost comes from the largest slice of an image pyramid. Inspired by this, we devise an anchor-based cascade framework, or anchor cascade, by introducing the anchors to cascade framework. More specifically, we try to embed a set of single-position anchors to a coarse image pyramid: (1) we use a set of anchors to handle a small range of object scales, e.g., from 12x12 pixels to 24x24 pixels, and thus avoid very large CNNs; (2) we use a coarse image pyramid to handle extremely large variations in object scale, and thus avoid the dense image pyramid. See a trade-off between a set of anchors and a dense image pyramid in Fig.~\ref{fig:coarse:pyramid}. As a result, anchor cascade significantly reduces the computational cost for dealing with large variations in object scale.

Although anchor cascade is computationally efficient, it still fails to recall some difficult faces such as tiny faces and those in profile. However, this can be addressed by exploring the contextual information, i.e., the areas surrounding the face region \cite{zhu2017cms,hu2016finding,Zhang_2017_ICCV}. Intuitively, for a fixed detection window, e.g., 24x24 pixels, additional context will always squeeze the face region. On the one hand, additional context such as face contours and ears are crucial for detecting low-resolution faces or faces in profile; On the other hand, face parts such as the eyes, nose, and mouth, help detect partially occluded faces \cite{yang2017faceness}. As a result, there will be a trade-off between adding additional context and preserving face region. Since the optimal context ratio for each candidate window is not available, we propose a maxout structure based on a diverse set of context templates. More specifically, for each candidate window, we use multiple different context templates, i.e., the context region increases with a fixed scale factor, and a candidate window is rejected only if all context templates fail to recall it. Considering that all context templates forms in a pyramid, we refer to this maxout structure as context pyramid maxout (CPM). As a result, we greatly improve the detection accuracy, especially on difficult faces. Specifically, the context pyramid maxout structure can be efficiently implemented using a parallel-style pipeline (see details in Section \ref{sec:cpm} and \ref{sec:imp}).

In summary, we have devised anchor cascade as well as a context pyramid maxout mechanism for efficient face detection: (1) to largely reduce the computational cost and (2) to significantly boost the detection accuracy. The anchor cascade face detector enables to train small models with a high detection accuracy: (1) it outperforms typical CNN-based cascade face detectors, e.g., MTCNN \cite{zhang2016joint}, with a large margin, while it still runs in comparable speed; and (2) it achieves comparable detection accuracy comparing with typical anchor-based face detectors, e.g., HR \cite{hu2016finding}, while it uses less than 1/25 parameters and 1/10 inference time. We conduct a number of experiments on two challenging face detection benchmarks, FDDB \cite{fddbTech} and WIDER FACE \cite{yang2016wider}, to demonstrate the effectiveness of the proposed anchor cascade framework.

%%%%%%%% Fig: main framework %%%%%%%%%%%%%
\begin{figure*}
\begin{center}
\includegraphics[width=1.0\linewidth]{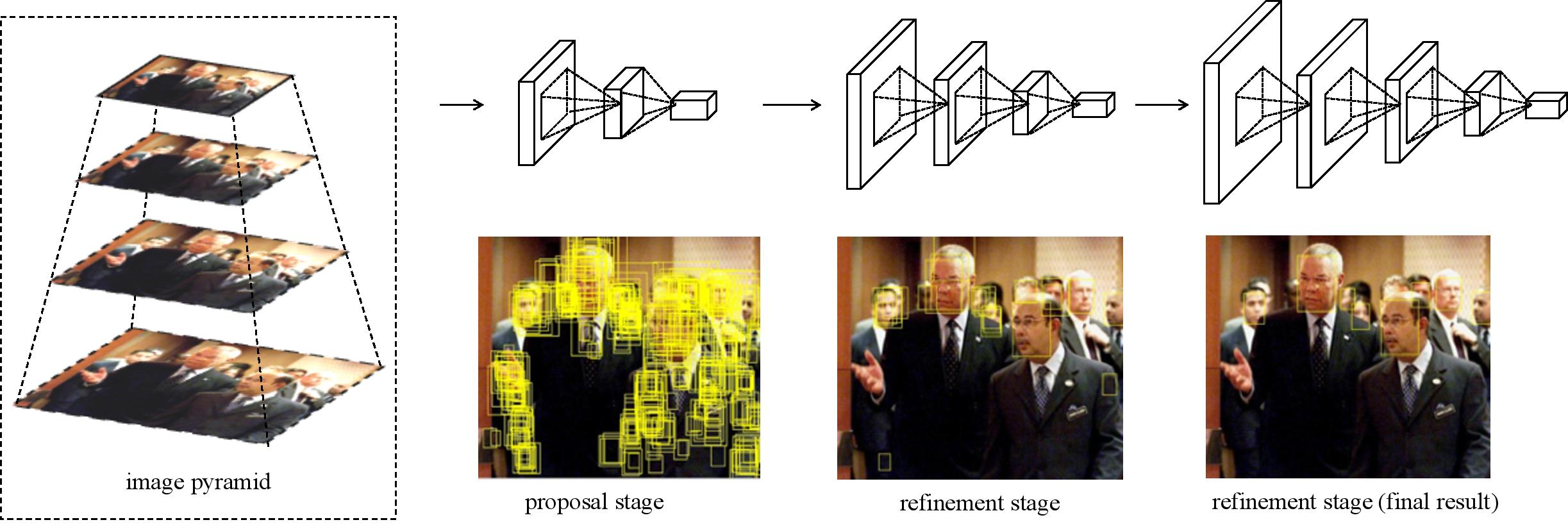}
\end{center}
   \caption{The main framework of CNN-based cascade face detector. It usually contains one proposal stage and several refinement stages. More specifically, the proposal stage usually is a FCN-based binary classifier, which generates a detection score map for each slice of the dense image pyramid. The refinement stages use larger CNNs to remove hard negative examples.}
\label{fig:framework}
\end{figure*}

%%%%%%%%%%%%% RELATED WORK %%%%%%%%%%%%
\section{Related Work}

\textbf{Cascade face detector}. 
The cascade face detector \cite{viola2004robust}, which uses Haar-like features and AdaBoost to train a cascade of binary classifiers, was the seminal work for real-time detection of near-frontal faces. The cascade face detector has since  been improved by using: (1) robust hand-crafted features, e.g., HOG \cite{zhu2006fast}, MB-LBP \cite{zhang2007face}, SURF \cite{Li_2013_CVPR}, ACF \cite{yang2014aggregate}, CCF \cite{yang2015convolutional}, NPD \cite{liao2016fast}, and LDF \cite{yuan2017robust}; (2) asymmetric feature selection approaches, e.g., FFS \cite{wu2008fast} and SAFS \cite{yu2016submodular}; (3) boosting algorithms, e.g., MILBoost \cite{zhang2006multiple}, RealBoost \cite{huang2007high}, and LACBoost \cite{shen2013training}; (4) new cascade structures, e.g., boosting chain \cite{xiao2003boosting}, Closed-Loop \cite{galteri2017spatio}, and soft-cascade \cite{bourdev2005robust}; and (5) multi-task learning, e.g., face/non-face classification, bounding box regression \cite{chen2014joint}, facial landmark localization \cite{zhang2016joint}, and face pose estimation \cite{zhu2012face, zhang2014improving}.

\textbf{CNN-based cascade face detector}. 
CNNs have recently contributed significantly to the progress in image classification \cite{krizhevsky2012imagenet}. Inspired by this, CNNs have been introduced to cascade face detection \cite{li2015convolutional} and improved by jointly training all stages \cite{qin2016joint}. Recently, CNN-based cascade face detectors have been further improved by using multi-task learning, i.e., jointly learning face/non-face classification, bounding box regression as well facial key point detection \cite{zhang2016joint}.

\textbf{Anchor-based face detector}.
Anchor-based detectors, which use multi-scale anchors at each sliding window position to simultaneously predict multiple different candidate regions, were first proposed in Faster R-CNN \cite{ren2015faster}. After that, SSD \cite{liu2016ssd} tried to assign different anchors to feature maps with different receptive fields. These anchor-based detectors have shown impressive performance in face detection \cite{jiang2017face}. Recently, anchor-based face detectors have been further improved by addressing (1) the mismatch between the receptive fields and anchor sizes \cite{zhang2017s3fd}, (2) the positions of facial landmarks \cite{chen2016supervised,li2016face}, and the scale distribution histogram of faces \cite{liu2017recurrent,hao2017scale}.

\textbf{Contextual information}. Contextual information has turned out to be crucial for object detection \cite{divvala2009empirical,li2018implicit} and recognition \cite{oliva2007role,garcia2017discriminant,song2017multi,lee2017going}. Recently, \cite{hu2016finding} analyzed the relationship between templates with different context regions. To explore contextual information in anchor-based face detectors, \cite{zhu2017cms} combined the features from different region of interests (RoIs), while \cite{najibi2017ssh} designed a context module using multiple different convolutional filters, e.g., 5x5 and 7x7 filters. Specifically, \cite{Ouyang_2017_ICCV} cascaded the predictions from different RoIs to efficiently boost the performance. To explore contextual information for cascade face detectors, \cite{Zhang_2017_ICCV} used an additional network to extract features with body information.

\section{Main Framework}
In this section, we first introduce the pipeline of CNN-based cascade face detector. We then describe the long-tail problem on computational cost and the anchor-based proposal network (APN) used in anchor cascade. After that, we introduce the context pyramid maxout structure, as well as its relationship with the multi-scale anchors. Finally, we briefly discuss the refinement stages, i.e., context-aware refinement networks.

\begin{figure*}[t]
\begin{center}
   \includegraphics[width=1.0\linewidth]{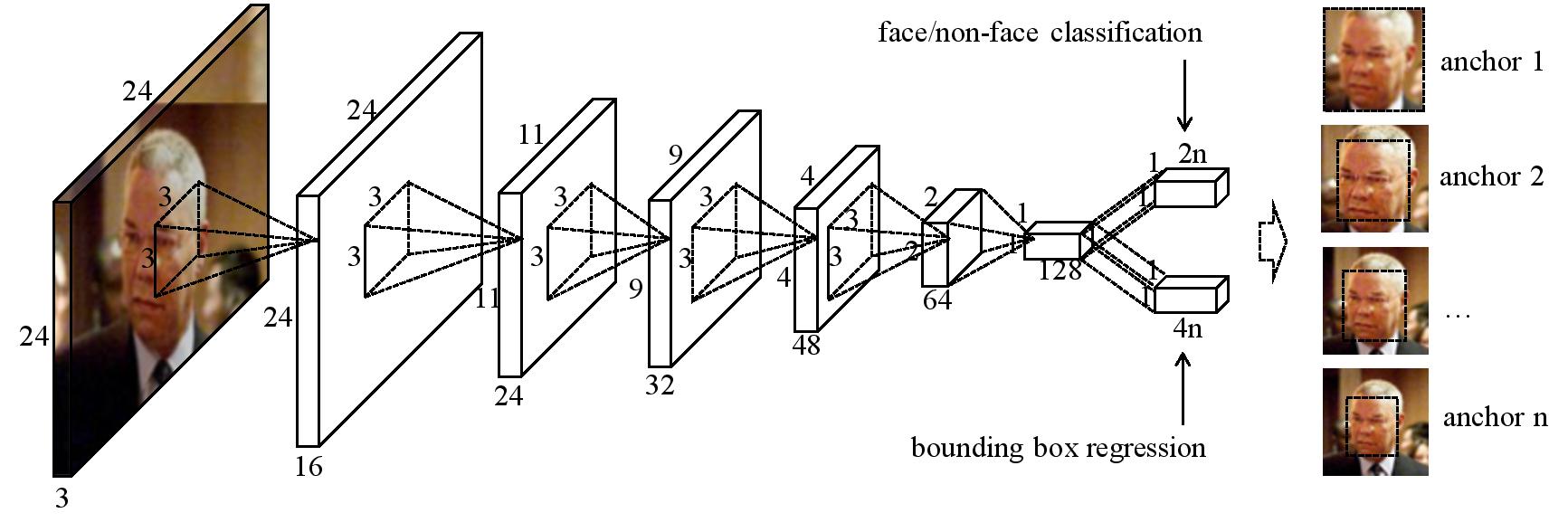}
\end{center}
   \caption{An example of anchor-based proposal network (APN). For training, APN uses 24x24 image patches as input to predict $n$ different proposals based on the same 128-$d$ features. During test, by replacing the last fully connected layer with a 1x1 convolutional layer, APN can detect image with arbitrary size.}
\label{fig:apn:detail}
\end{figure*} 

\subsection{CNN-based Cascade Face Detector}
%CNN-based cascade face detector usually contains two or three stages: one proposal stage, and several refinement stages. 
Cascade face detectors rely on a dense image pyramid to detect faces of different scales, i.e., it slides a detection window and detects single-scale faces on each slice of the dense image pyramid. Inspired by \cite{sermanet2014overfeat}, the fully convolutional neural networks \cite{long2015fully} (FCN)-based sliding-window approach has been widely used in CNN-based cascade face detectors \cite{qin2016joint,zhang2016joint,Zhang_2017_ICCV}. More specifically, by replacing last fully connected layer with a 1x1 convolutional layer, the FCN-based sliding window approach shares the computation of the overlapped area between different windows.

\textbf{Image pyramid}. Given an image, a dense image pyramid is built by resizing the original image according to the size of detection window. For example, given the size of detection window $S_w$, if we want to detect faces of scale $S$, the original image then need to be resized by $S_w/S$. In practice, we usually construct the image pyramid using a fixed scale factor $\alpha_I$, e.g., $0.7937$. That is, if the first slice of image pyramid is the original image, the $k$-th slice then is obtained by resizing the original image to $\alpha_I^{k-1}$.

\textbf{Proposal generation}. During test, by replacing the last fully connected layer with a 1x1 convolutional layer, the proposal network can deal with arbitrary input size and a score map will be generated for the input. After that, all candidate windows can be decoded from the score map and a majority of non-face windows are removed by a score threshold as well as the non-maximum suppression (NMS).

\textbf{Refinement}. In candidate windows generated by the proposal stage, a number of non-face windows still exist. As a result, refinement stages usually train larger networks by mining hard examples from the proposal stage. To remove non-face windows, one or two refinement stages are generally required.

\begin{figure}[t]
\begin{center}
   \includegraphics[width=1.0\linewidth]{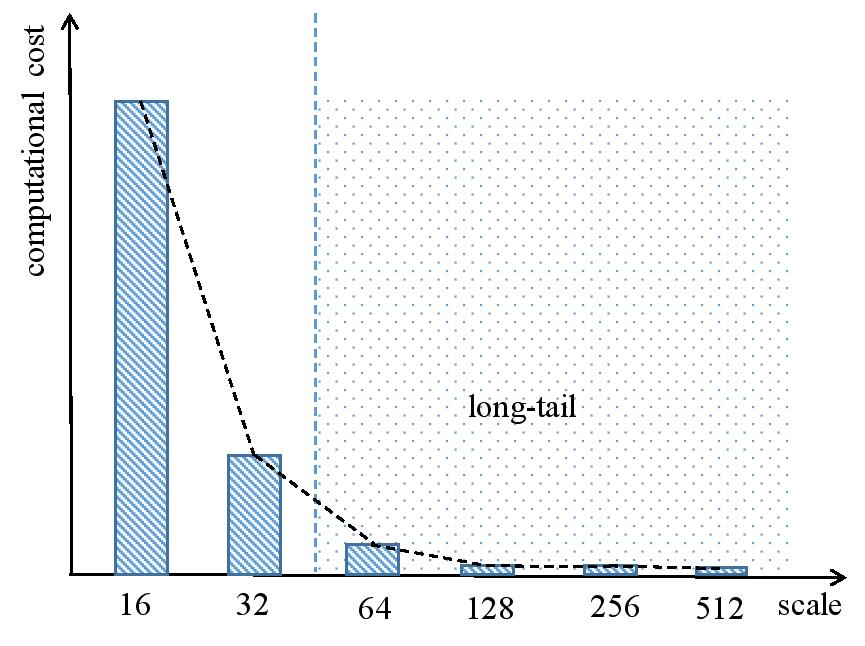}
\end{center}
   \caption{An example of long-tail problem on computational cost. If a single-scale face detector scans 32x32-pixels faces in the original image, it then needs to resize the original image by a factor of $2$ to detect 16x16-pixels faces. Furthermore, the computational cost will become four times. As a result, to detect faces from 16x16 to 512x512 pixels, the computational cost in the shaded area, i.e., from 64x64 pixels to 512x512 pixels, is trivial. We refer to this problem as the long-tail property on computational cost.}
\label{fig:longtail:demo}
\end{figure}

\subsection{Anchor-based Proposal Network}

To design an efficient proposal network, we first describe the long-tail problem on computational cost for scanning a dense image pyramid, e.g., in cascade face detectors. Specifically, we find that the majority of computational cost comes from detecting objects within a relatively small range of object scales, e.g., from 16x16 pixels to 32x32 pixels. As a result, we refer to this problem as the long-tail problem on computational cost and we given an example of the long-tail problem in Fig. \ref{fig:longtail:demo}. 

Multi-scale anchors share the computations for detecting objects of different scales. However, the long-tail problem on computational cost has not been well addressed by anchor-based detectors for efficient face detection. More specifically, anchor-based detectors usually use multi-scale anchors to cover an extremely wide range of object scales, e.g., from 16x16 to 512x512 pixels. As a result, it heavily relies on a very large base network to learn scale-invariant features. We argue that it is more computationally efficient to deal with an extremely large variations in object scale by using a coarse image pyramid. To avoid very large CNNs by addressing the long-tail problem, we thus design our anchor-based proposal network according to the following rules: (1) we use a set of anchors to cover only a small range of object scales, e.g, from 16x16 pixels to 32x32 pixels; (2) we use a coarse image pyramid to cover the long-tail areas in Fig. \ref{fig:longtail:demo}. 

\begin{figure*}[!t]
\begin{center}
   \includegraphics[width=1.0\linewidth]{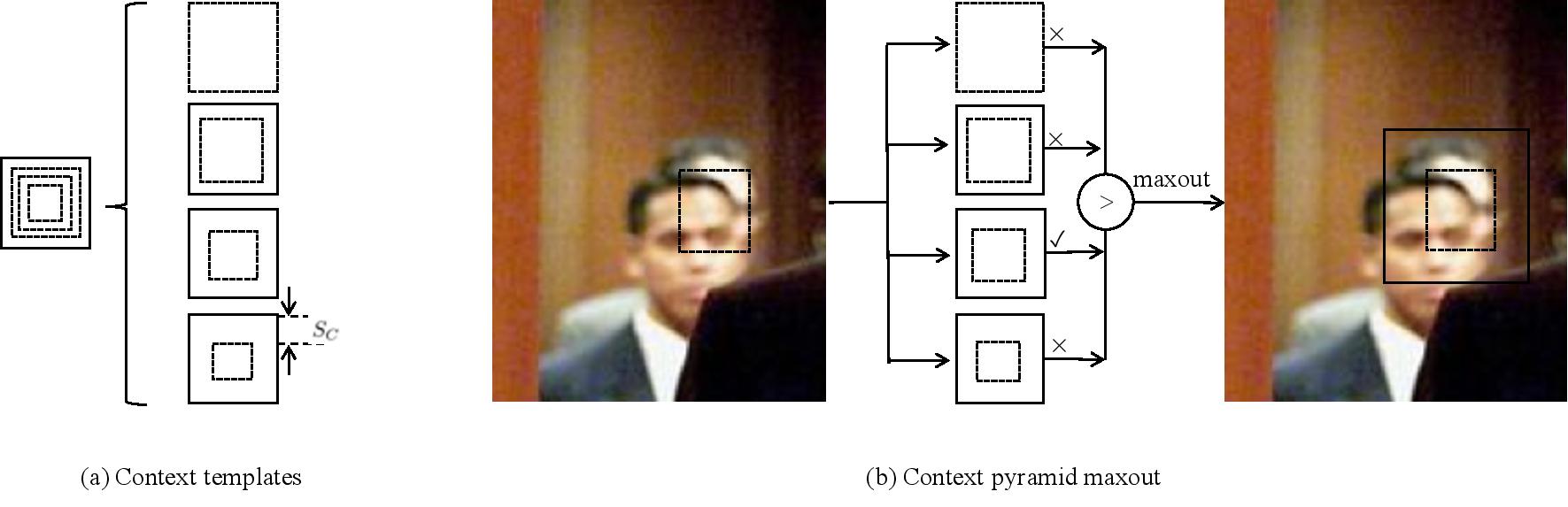}
\end{center}
   \caption{An example for context pyramid maxout (CPM). In Fig. (a), it is a set of four context templates with different context regions. In Fig. (b), for each candidate window (dashline area), we use multiple context templates and only the template with maximum prediction score are kept for further processing.}
\label{fig:cpm:demo}
\end{figure*} 

We describe the details of our anchor-based proposal network (APN) for efficient face detection as follows. In real-world scenarios, it is  computationally intensive to detect small or tiny faces e.g., 20x20 pixels in FDDB and 10x10 pixels in WIDER FACE. Inspired by this, we use a detection window 24x24 pixels with a set of anchors as follows: given the size of detection window $S_W = 24$, the size of $k$-th anchor is defined by
\begin{equation}
	S_A(k) = S_W * \alpha_A^{k-1},
\end{equation}
where $\alpha_A$ is a scale factor to control the density of anchors. In this paper, we always use $\alpha_A=\alpha_I$ and a set of anchors thus coincide with an image pyramid. That is, while the largest anchor of APN works on the $k$-th slice of an image pyramid, other smaller anchors will equivalently work on the $(k+1)$-th, $(k+2)$-th, \dots, slices of the image pyramid. As a result, APN simultaneously scans multiple slices of a dense image pyramid, and thus only needs a coarse image pyramid, i.e., $1/n_A$ of all slices in the original dense image pyramid, where $n_A$ is the number of anchors in APN.  For example, given $\alpha_A=0.7937$ and $n_A=4$, the scale factor of the coarse image pyramid reduces from $0.7937$ to $0.7937^{4} \approx 0.4$. Considering that the long-tail problem, the computational cost will further reduce to less than $1/n_A$, i.e.,
\begin{equation}
\frac{C_{n_A}}{\sum_{k=1}^{n_A} C_{k}} = \frac{\alpha^{n_A}}{\sum_{k=1}^{n_A} \alpha^{k}} = \frac{\alpha^{n_A-1}(1-\alpha)}{1-\alpha^{n_A+1}},
\end{equation}
where $\alpha=\alpha_A^2$ and $C_k$ denotes the computational cost on the $k$-th slice of an image pyramid. As a result, for $\alpha_A=0.7937$ and $n_A=4$, the computational cost will reduce to approximately $1/9.74$.

Anchors used in APN are inspired by the region proposal network (RPN) used in Faster R-CNN \cite{ren2015faster}, the aim being to share the computations of the dense image pyramid. However, there are several differences between APN and RPN. For simplicity, we refer to these anchors as APN anchors and RPN anchors, respectively. The differences then can be described as follows. First, APN uses a small base network, which does not need to be pre-trained on ImageNet; Second, RPN allocates anchors on a $W \times H$ feature map, while APN contains only a set of single-position anchors and is trained using image patches as a simple classification network; Third, APN anchors only cover a small range of object scales and thus do not suffer from the problem of a mismatch between anchors and the receptive fields \cite{cai2016unified,zhang2017s3fd}.

%%% Context Pyramid Maxout %%%%
\subsection{Context Pyramid Maxout}
\label{sec:cpm}

Contextual information has turned out to be crucial for face detection, especially for detecting challenging faces such as tiny faces \cite{zhu2017cms,hu2016finding,Zhang_2017_ICCV}. However, for CNN-based cascade face detectors, there is always a trade-off between adding additional context and preserving face regions due to the fixed detection window. See Fig. \ref{fig:cpm:demo} (a) for a set of context templates with different context regions. For each candidate window, the optimal context region varies due to the large variations in illumination, occlusion, and pose. That is, we do not have prior information on context template selection. To improve the recall rate by using proper context templates, we use a diverse set of context templates with a maxout structure. More specifically, given a candidate window $x$, let $p_i(x)$ denote the score given by the $i$-th context template, the maxout score then is evaluated as follows:
\begin{equation}
	p(x) = \max\limits_{i=1, 2, \dots, n_C} p_i(x),
\end{equation}
where $n_C$ is the number of context templates. By using the maxout score, we will only reject a candidate window if all context templates fail to recall it. A diverse set of context templates can be constructed as follows. Let $S_C(i)$ denote the padding size of the $i$-th context template, we then have
\begin{equation}
	S_C(i) = \frac{S_W*(1-\alpha_C^{i-1})}{2},
\end{equation}
where $\alpha_C \in (0,1]$ is a scale factor. Furthermore, a set of context templates can be efficiently implemented using the same structure with a set of anchors. See more details about the implementation of context pyramid maxout in Section \ref{sec:imp}.

\begin{figure*}[t]
\begin{center}
   \includegraphics[width=0.95\linewidth]{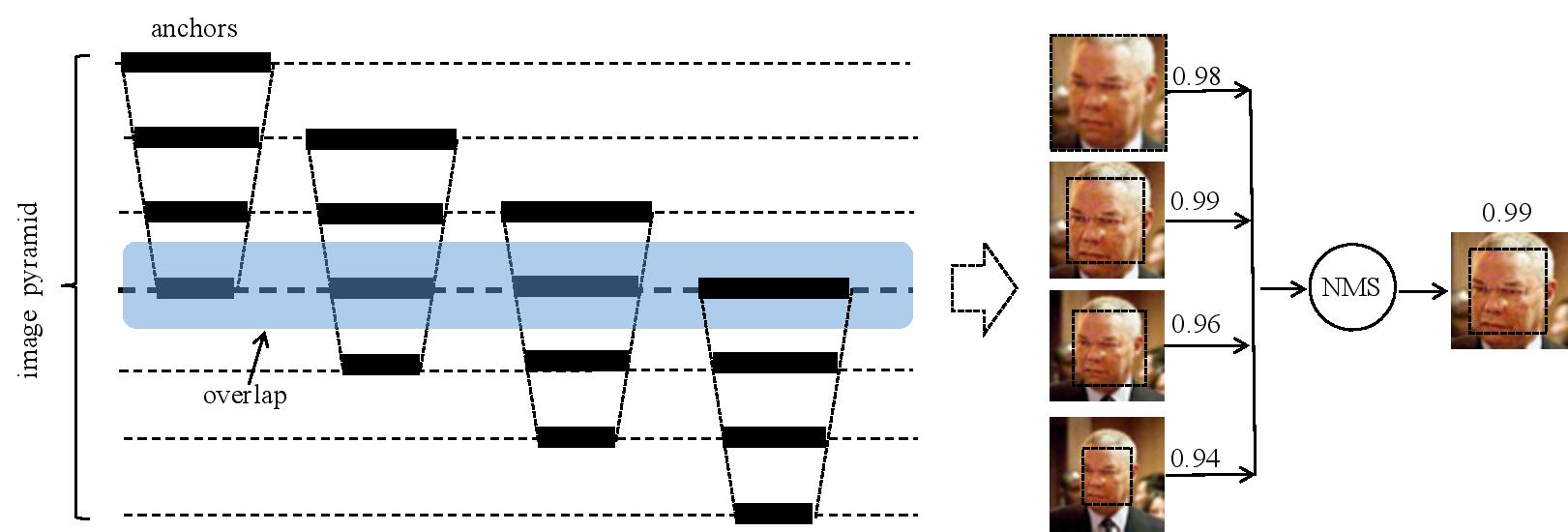}
\end{center}
   \caption{An efficient implementation of context pyramid maxout. Given an APN with $n_A$ anchors, a context pyramid on the $k$-th slice of the image pyramid is constructed as follows: the $i$-th context templates is the $i$-th anchor of APN, i.e., APN actually works on the $(k-i+1)$-th slice of the image pyramid. As a result, a set of context templates exist on the $k$-th slice of the image pyramid constructed by APN working on the $k$-th, $(k-1)$-th, \dots, slices of the image pyramid; that is, the overlap between APN anchors forms a set of context templates in a parallel-style pipeline. The maxout structure can be trivially implemented by using NMS with a large threshold, e.g., 0.9.}
\label{fig:cpm:detail}
\end{figure*}

%%%%%% Context Aware Refinement.%%%%%%%%%%%
\subsection{Context-Aware Refinement}

The motivation of each refinement stage is to reject as many non-face windows as possible and simultaneously retain any face windows. The refinement stage thus is important to boost the performance of a face detector on hard faces. More specifically, our refinement networks share the same structure with APN for training except that (1) refinement networks have more layers and larger input size, e.g., 48x48 pixels for RNet48 and 96x96 pixels for RNet96; (2) refinement networks are trained by mining the hard examples from the proposal stage. 

During testing, we perform context-aware refinement as follows. Given each candidate window retained by the proposal stage, we also obtain a context template in maxout structure, i.e., the ``optimal" template for candidate window $x$ is
\begin{equation}
	i^* \in \underset{i=1, 2, \dots, n_C}{\arg\max} \enskip p_i(x).
\end{equation}
After that, we use the $i^*$-th context template of the refinement network to classify the candidate window $x$. Considering all stages in our cascade framework are based on anchors, we refer to the proposed framework as anchor-based cascade or anchor cascade.

%%%%%% Inplementation Details %%%%%%%%%%%
\section{Implementation Details}
\label{sec:imp}
In this section, we first describe the implementation details of context pyramid maxout. We then introduce the multi-task loss as well as the network architectures used in each stage of anchor cascade face detector. Finally, we discuss data preparation as well as hyper parameters for training.

\subsection{Implementation of CPM}
To share the computations between different context templates, we find that a set of anchors share the same input in APN, and each anchor thus contains different context information. Inspired by this, our context templates share the same scale factor with APN, i.e., $\alpha_C=\alpha_A$, and a set of anchors thus form a diverse set of context templates (see Fig. \ref{fig:cpm:demo} (a) for an example). More specifically, the context templates induced by the multi-scale anchors run in a parallel pipeline style. We demonstrate the parallel pipeline of context pyramid maxout in Fig. \ref{fig:cpm:detail}. As a result, the maximum number of context templates $n_C$ is limited by the number of anchors in APN, i.e., 
\begin{equation}
n_C = n_A-i+1.	
\end{equation}
A shortcoming for such an implementation is that more context templates always degrade the computational efficiency of APN. That is, there will be a trade-off between using more context templates and keeping computational efficiency.

\subsection{Multi-task Loss}
Face/non-face classification and bounding box regression are two important tasks in object detection. It has turned out that joint learning classification and regression tasks improves object detectors \cite{ren2015faster, zhang2016joint, yu2016unitbox, ranjan2016hyperface}. Inspired by this, we train our anchor cascade in a similar multi-task learning framework \cite{zhang2016joint}. For face/non-face classification, we use the softmax loss, i.e., the softmax activation function with cross entropy loss. For bounding box regression, we adopt the Euclidean loss based on the parameterizations of four coordinates as follows. Let $x_a^1, y_a^1, x_a^2, y_a^2$ denote the coordinates of the top-left corner and bottom-right corner of the candidate window, and let $x_g^1, y_g^1, x_g^2, y_g^2$ denote the ground truth. The regression parameters then are defined as follows:
\begin{align}
	\delta_x^1 &= (x_g^1 - x_a^1) / w, \\ 
	\delta_y^1 &= (y_g^1 - y_a^1) / h, \\
	\delta_x^2 &= (x_g^2 - x_a^2) / w, \\
	\delta_y^2 &= (y_g^2 - y_a^2) / h,
\end{align}
where $w$ and $h$ are the width and height of the candidate window, respectively. Specifically, all context templates share the same regression parameters, i.e., the regression parameters are evaluated without considering the padding size. The final loss function uses a loss weight $\lambda$ to balance the face/non-face classification loss $L_{cls}$ and bounding box regression loss $L_{reg}$, i.e.,
\begin{equation}
	L_{total} = L_{cls} + \lambda * L_{reg}.
\end{equation}

\subsection{Network Architectures}
\begin{figure*}[!t]
\begin{center}
   \includegraphics[width=1.0\linewidth]{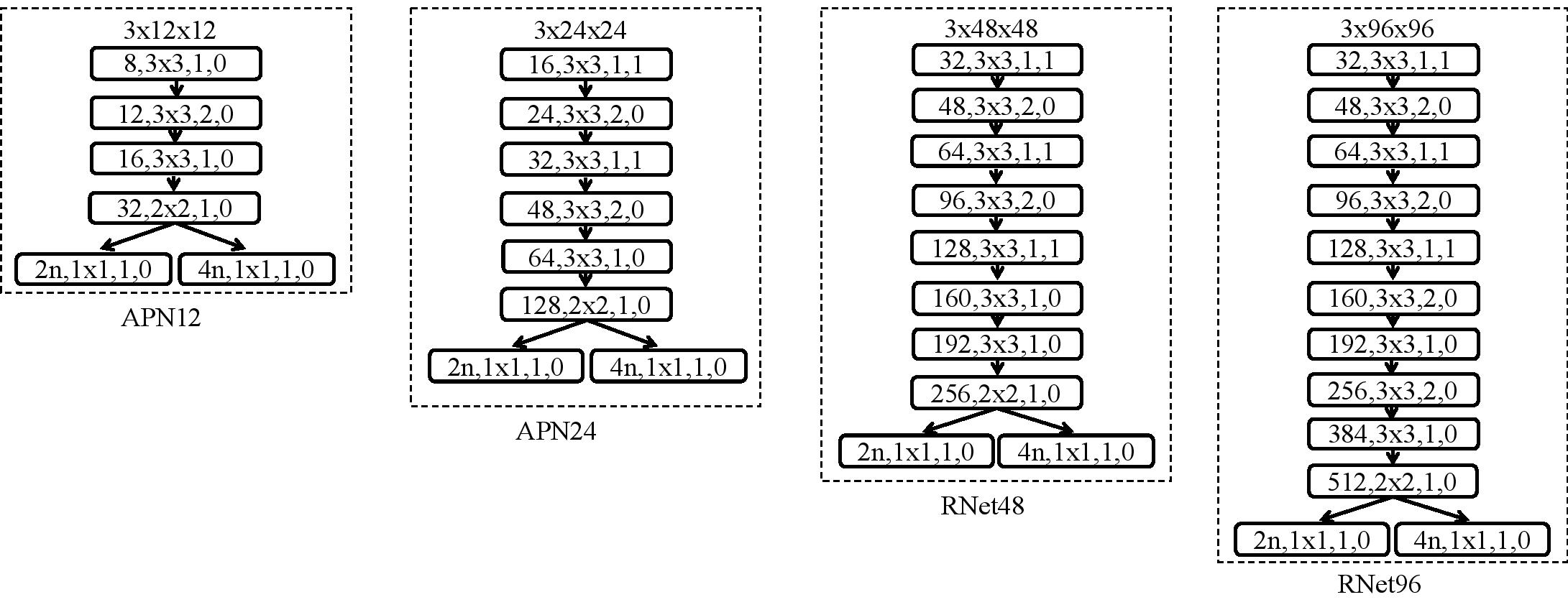}
\end{center}
   \caption{A typical configuration of network architectures used in anchor cascade. The text in convolutional layers refers to the number of channels, size of kernel, stride, and padding size, respectively. We use four anchors $n_A=4$ in our current implementation, i.e., $n=4$.}
\label{fig:network}
\end{figure*}

For simplicity,  we use plain CNNs, i.e., stacked convolutional layers, in both the proposal and refinement stages of the anchor cascade face detector. Unlike \cite{li2015convolutional,zhang2016joint,qin2016joint}, we use convolutional layers with stride 2 to replace all max-pooling layers. As a result, all networks used in anchor cascade face detector contain only 3x3 convolutional layers followed by a PReLU \cite{he2015delving} activation function, except the last two layers (see a typical configuration of network architectures in Figure \ref{fig:network}).

\subsection{Data Preparation and Training}
We collect training and validation data from the training set of WIDER FACE \cite{yang2016wider}. Inspired by \cite{li2015convolutional,zhang2016joint}, we randomly generate positive (IoU $>$ 0.65) and negative (IoU $<$ 0.3) samples by keeping a ratio pos:neg=1:3. For bounding box regression, we randomly collect semi-positive (IoU $>$ 0.4) samples with the same ratio as positive samples. For the proposal stage, we use 0.8M positive samples. For the two refinement stages, we use 1M and 1.1M positive samples, respectively. More specifically, in the first refinement stage, we use 3.0M negative samples (1.0M from the proposal stage; 2.0M hard negative samples). In the second refinement stage, we use 3.3M negative samples (1.1M from the first refinement stage; 2.2M hard negative samples).

The proposed anchor cascade is implemented using a customized Caffe \cite{jia2014caffe}. We use $\lambda=1.0$ for the second refinement stage and $\lambda=0.5$ for other stages. All models are trained using stochastic gradient descent (SGD) with momentum 0.9 and weight decay 2e-5. For the proposal stage, we start with a learning rate 0.1 for 4 epochs and then divide by 5 for every 2 epochs, with 12 epochs in total. For both refinement stages, we use a learning rate of 0.01 for 5 epochs and 0.001 for 3 epochs. We use the batch size 480, 360 and 240 for APN24, RNet48, and RNet96, respectively. Specifically, we use a dropout of 0.5 for the second refinement stage, i.e., RNet96.

%%%%%% EXPERIMENT %%%%%%%%%%%%%%%%
\section{Experiment}

In this section, we first briefly introduce two widely used benchmarks in face detection, FDDB \cite{fddbTech} and WIDER FACE \cite{yang2016wider}. We then conduct a number of experiments to demonstrate the effectiveness of the proposed anchor cascade framework.

\subsection{Datasets and Benchmarks}

\textbf{FDDB} contains 2,845 images with 5,171 faces. For FDDB evaluations, we follow the ``unrestricted training" setting, i.e., we use all ten folds of the dataset as validation data without performing 10-fold cross-validation.

\textbf{WIDER FACE} contains 32,203 images with 393,793 faces, of which $40\%$ are used for training, $10\%$ for validation, and $50\%$ for testing. According to the detection rate, the validation data are divided into three classes: ``easy", ``medium", and ``hard". We train all our models using the training set and test on the validation set.

\begin{figure}
\includegraphics[width=1.0\linewidth]{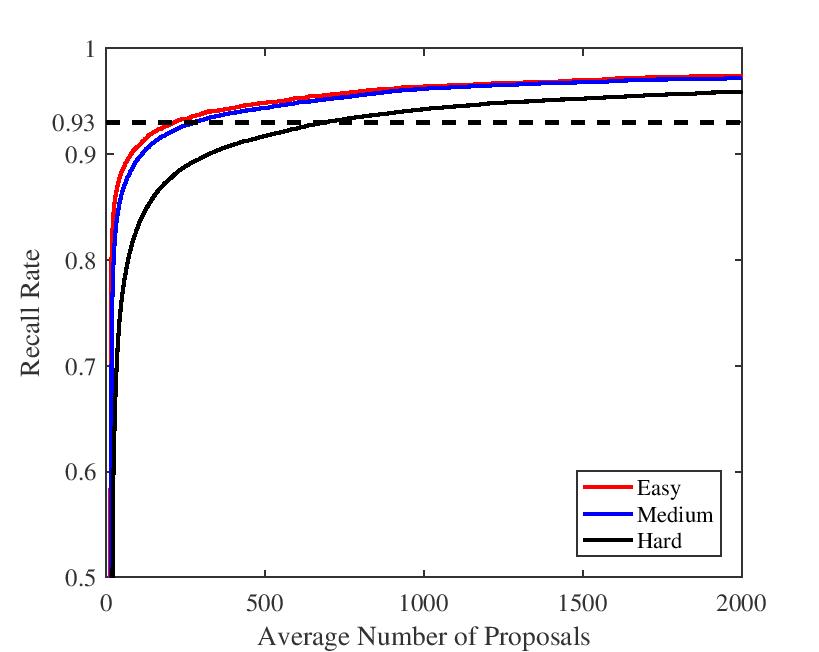}
\caption{The recall rate of APN24 on WIDER FACE. Specifically, we use min-face $S_{min}=10$ and $n_C=3$ in this experiment.}
\label{fig:exp:recall:wider}
\end{figure}

\begin{figure*}[t]
\begin{center}
	\includegraphics[width=1.0\linewidth]{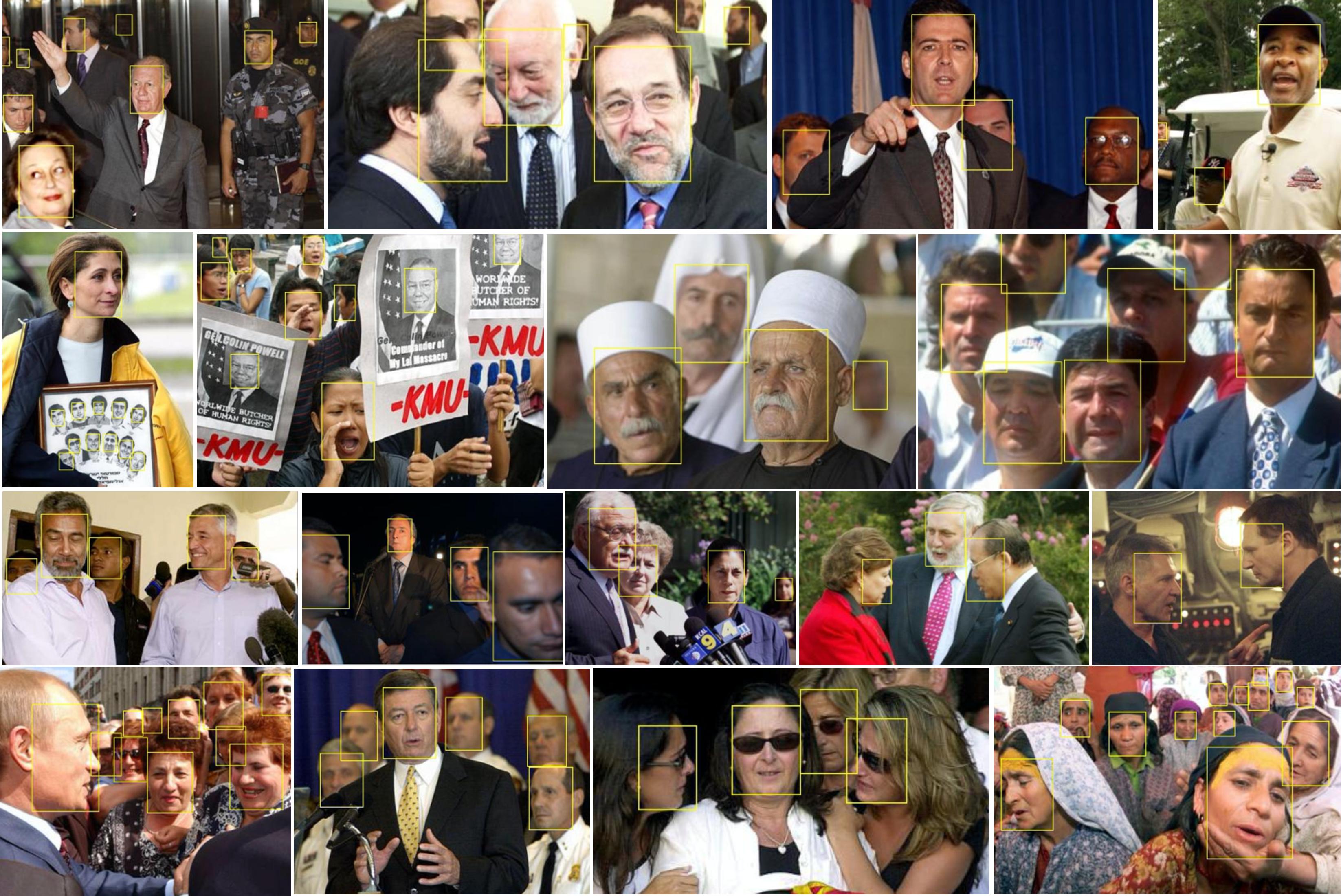}
\end{center}
\caption{Qualitative results on FDDB and best view in color.}
\label{fig:exp:fddb:demo}
\end{figure*}

\subsection{Effectiveness for Proposal Generation}

\begin{table*}
\caption{The recall rate of APN24 on FDDB. We use min-face $S_{min}=16, 20, 24$, as FDDB dataset contains face around 20x20 pixels. Specifically, APN24(S) shares the same structure with APN24, except the number of channels in each convolutional layer, i.e., 8, 16, 24, 32, 64, 128, respectively.}
\label{tab:exp:recall:fddb}
\begin{center}
\begin{tabular}{|l|c|c|c|c|r|r|l|c|c|c|c|r|r|}
\hline
Network & $n_C$ & $S_{min}$ & R@50 & R@100 & Model & Speed & Network & $n_C$ & $S_{min}$ & R@50 & R@100 & Model & Speed \\
\hline
PNet \cite{zhang2016joint} & 1 & 16 & 0.9604 & 0.9712 &  28KB & 25.17 FPS & APN24 & 3 & 16 & 0.9920 & 0.9948 & 360KB & 15.73 FPS\\
\hline
PNet \cite{zhang2016joint}  & 1 & 20 & 0.9557 & 0.9700 & 28KB & 33.75 FPS & APN24 & 3 & 20 & 0.9898 & 0.9919 & 360KB & 23.16 FPS \\
\hline
PNet \cite{zhang2016joint}  & 1 & 24 & 0.9540 & 0.9673 & 28KB & 43.15 FPS & APN24 & 3 & 24 & 0.9851 & 0.9876 & 360KB & 30.13 FPS\\
\hline
\hline
APN24 & 1 & 16 & 0.9872 & 0.9910 & 360KB & 22.91 FPS & APN12 & 2 & 16 & 0.9598 & 0.9685 & 69KB & 43.31 FPS \\
\hline
APN24 & 1 & 20 & 0.9860 & 0.9882 & 360KB & 32.63 FPS & APN12 & 2 & 20 & 0.9484 & 0.9553 & 69KB & 55.78 FPS \\
\hline
APN24 & 1 & 24 & 0.9825 & 0.9856 & 360KB & 43.39 FPS & APN12 & 2 & 24 & 0.9354 & 0.9408 & 69KB & 72.11 FPS \\
\hline
\hline
APN24 & 2 & 16 & 0.9907 & 0.9946 & 360KB & 20.82 FPS & APN24(S) & 2 & 16 & 0.9760 & 0.9885 & 164KB & 38.33 FPS\\
\hline
APN24 & 2 & 20 & 0.9888 & 0.9921 & 360KB & 29.68 FPS & APN24(S) & 2 & 20 & 0.9604 & 0.9743 & 164KB & 51.41 FPS \\
\hline
APN24 & 2 & 24 & 0.9859 & 0.9901 & 360KB & 38.02 FPS & APN24(S) & 2 & 24 & 0.9564 & 0.9624 & 164KB & 65.81 FPS  \\
\hline
\end{tabular}
\end{center}
\end{table*}

\begin{figure*}[!t]
\begin{center}
	\includegraphics[width=1.0\linewidth]{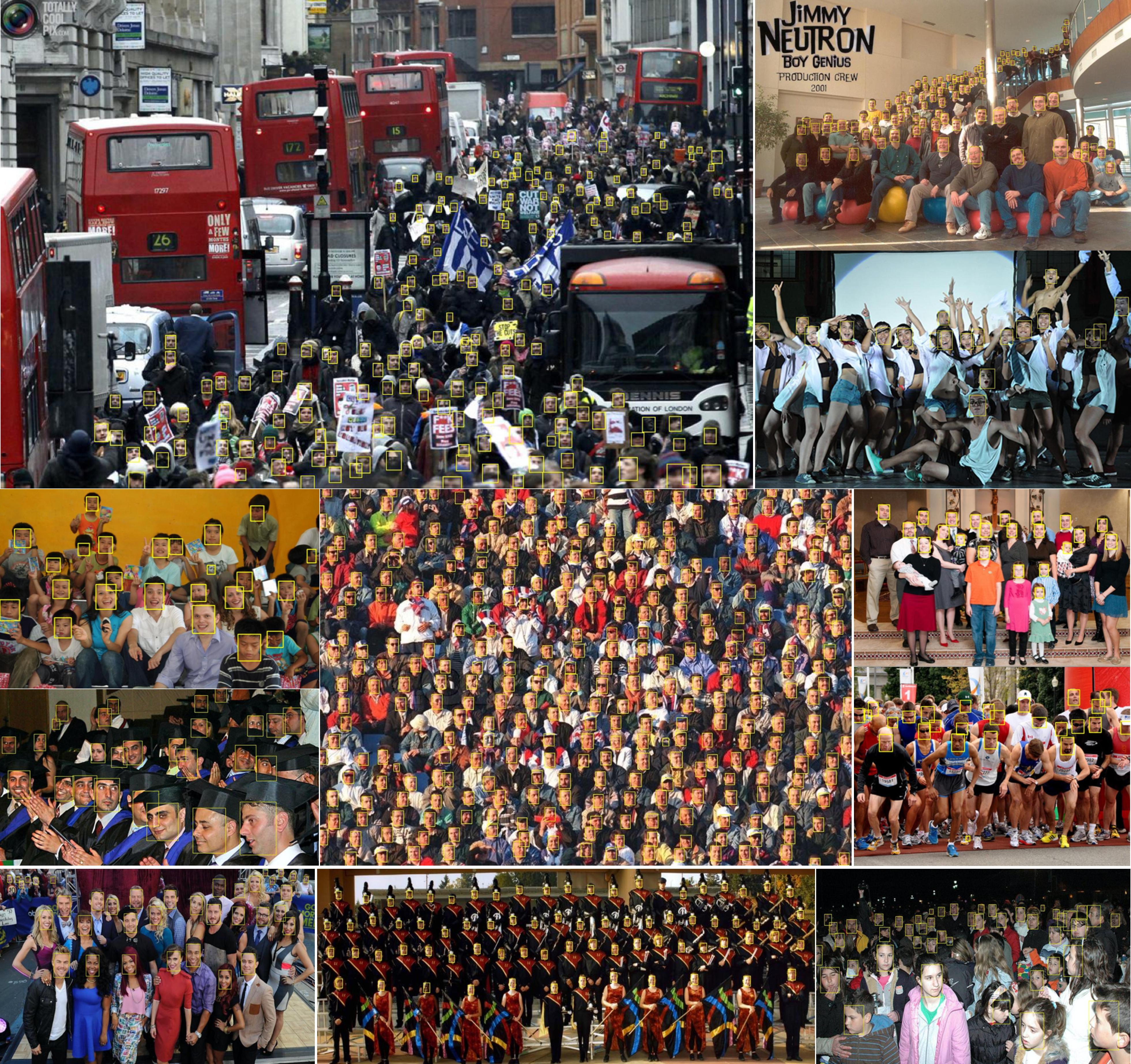}
\end{center}
\caption{Qualitative results on WIDER FACE. Please zoom in to see tiny detections.}
\label{fig:exp:demo}
\end{figure*}

Proposal generation is very important for efficient face detection, especially for cascade face detectors. On the one hand, refinement stages are based on generated proposals and they will never recall any faces outside generated proposals; On the other hand, refinement stages usually contain larger networks for hard examples and thus are not efficient for thousands of proposals. As a result, it is crucial for efficient face detection to generate high quality proposals.

In this experiment, we train several APNs using input size 12x12 and 24x24 pixels, i.e., APN12 and APN24, respectively. To demonstrate effectiveness of APNs for proposal generation, we compare the recall rate of APNs under different number of proposals. More specifically, in Table \ref{tab:exp:recall:fddb}, we evaluate the recall rate on FDDB using 50 and 100 proposals per image (in average). We find that (1) APN24 achieves very high recall rate comparing with PNet used in MTCNN, e.g., $0.9825 \textbf{ vs } 0.9540$; (2) APN24 runs in comparable speed with PNet, even though it is 12 times larger than PNet. Furthermore, with an average of 250 proposals per image, APN24 recalls almost all faces in FDDB, i.e., more than $99.9\%$ faces. To further demonstrate the efficiency of APNs, we train two small models, i.e., APN12 and APN24(S). We find that both APN12 and APN24(S) run very fast, and APN24(S) outperforms PNet with a clear margin. A possible reason for the weakness of APN12 is that the smallest anchor in APN12, i.e., 6x6 pixels, is too small to carry useful information for challenge faces. In Figure \ref{fig:exp:recall:wider}, we demonstrate the recall rate on WIDER FACE, which contains a large number of tiny faces. We see that APN24 recalls more than $93\%$ faces with less than 1k proposals per image on ``hard" set of WIDER FACE. By comparison, the default face detector used in WIDER FACE, i.e., Faceness-Net \cite{yang2015facial}, achieves a recall rate $93.1\%$ with 10k proposals.

\subsection{Two-stage Anchor Cascade}

APNs achieve very high recall rate, while there are still a large number of false positives in generated proposals. To further remove hard false positives, we use a refinement network RNet48 and demonstrate its effectiveness for efficient face detection by comparing with (1) a popular CNN-based cascade face detector MTCNN \cite{zhang2016joint}; and (2) a typical anchor-based face detector HR \cite{hu2016finding}. In Table \ref{tab:exp:rnet48:fddb}, we see that (1) the two-stage anchor cascade face detector is comparable with MTCNN in both model size and running speed, while it greatly outperforms MTCNN in detection accuracy; (2) it achieves comparable performance with HR on FDDB using less than $1/25$ parameters and $1/10$ inference time. As a result, with anchor cascade, we further bridge the gap between CNN-based cascade face detectors and anchor-based face detectors. 

\begin{table}
\caption{The performance of the two-stage anchor cascade on FDDB. Specifically, we use min-face $S_{min}=18$ for both APN24(S) and APN24.}
\label{tab:exp:rnet48:fddb}
\begin{center}
\begin{tabular}{|l|c|c|r|r|}
\hline
Method & $n_C$ & FP=1k & Model & Speed  \\
\hline
MTCNN \cite{zhang2016joint} & - & 0.9435 & 2.10MB & 25.09 FPS\\
HR-ResNet101 \cite{hu2016finding} & - & 0.9698 & 103.7MB & 2.12 FPS\\
\hline
\hline
APN24(S)+RNet48 & 1 & 0.9416 & 3.56MB & 40.79 FPS\\
APN24+RNet48 & 1 & 0.9704 &  3.76MB & 24.74 FPS\\
\hline
\end{tabular}
\end{center}
\end{table}

\begin{figure*}[!t]
\centering
\subfloat[Easy]{\includegraphics[width=2.5in]{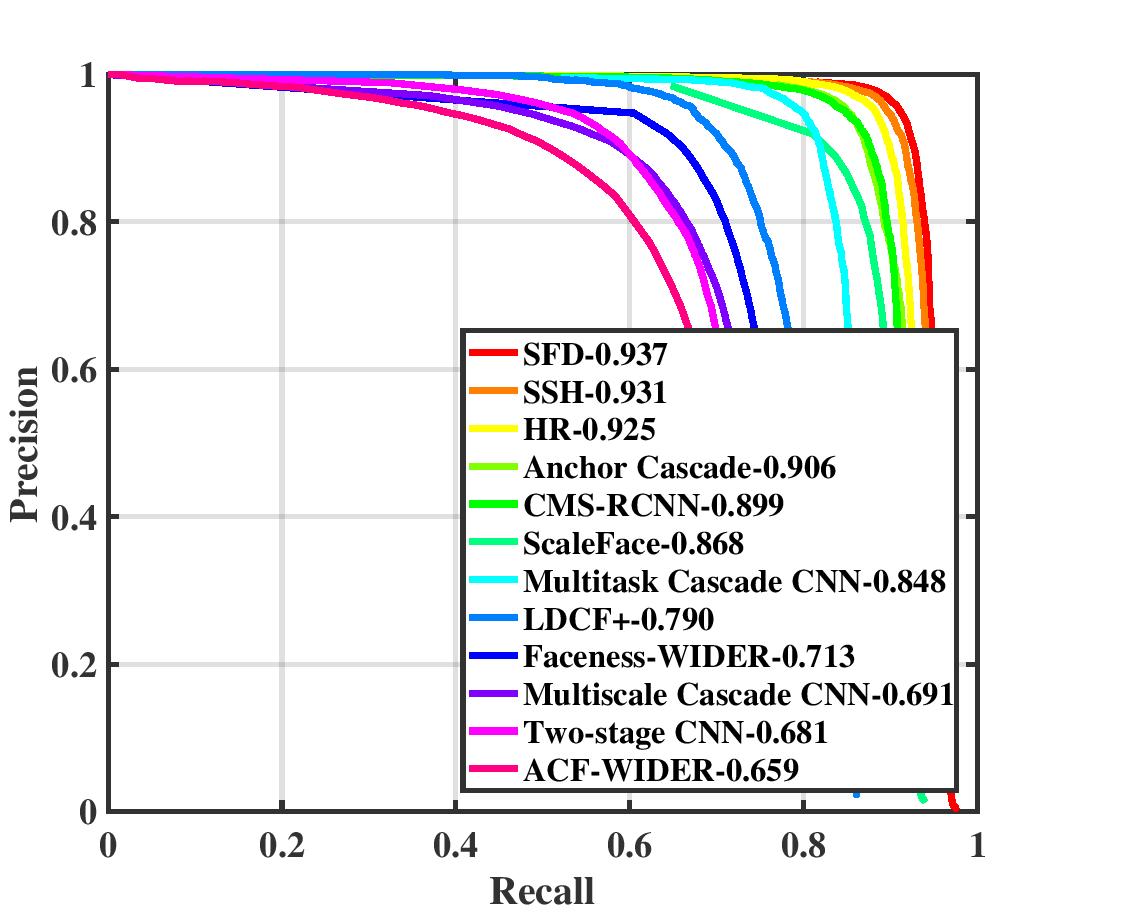}%
\label{fig:exp:wider:easy}}
\subfloat[Medium]{\includegraphics[width=2.5in]{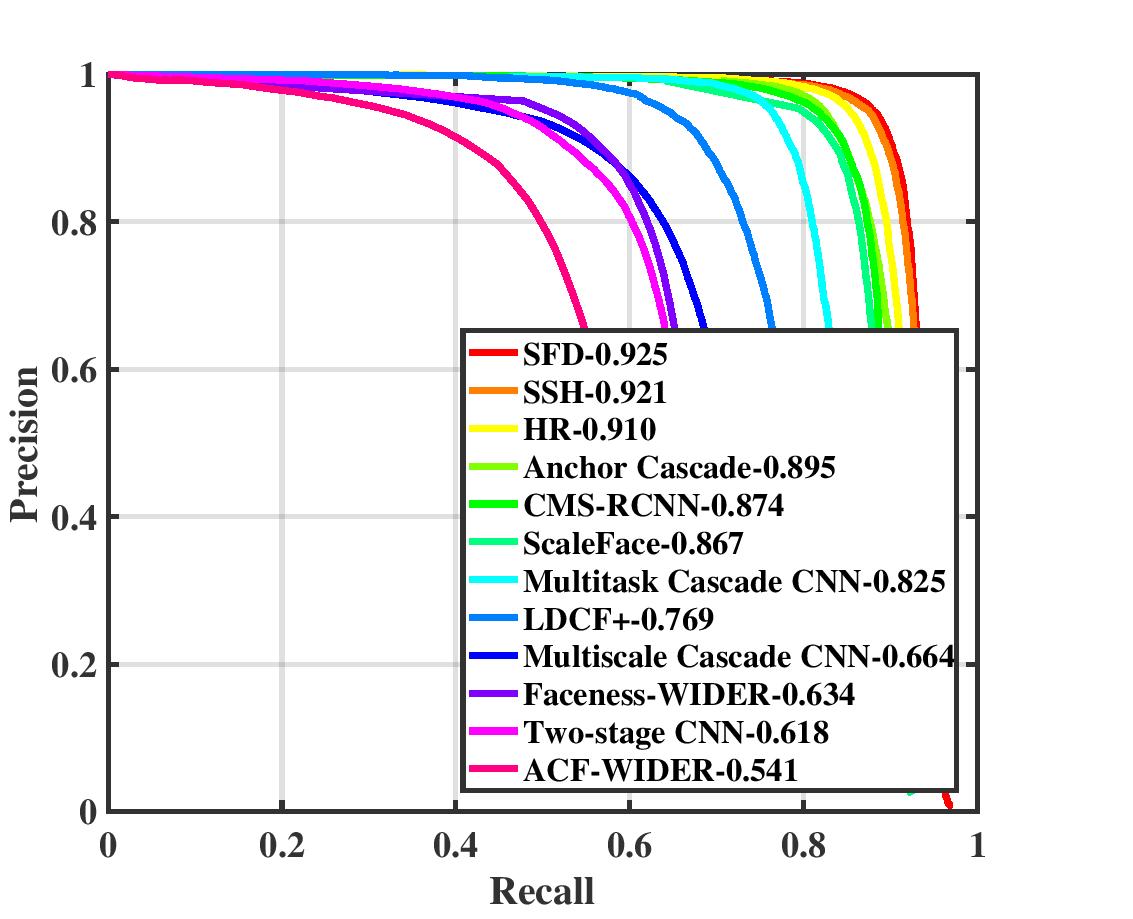}%
\label{fig:exp:wider:medium}}
\subfloat[Hard]{\includegraphics[width=2.5in]{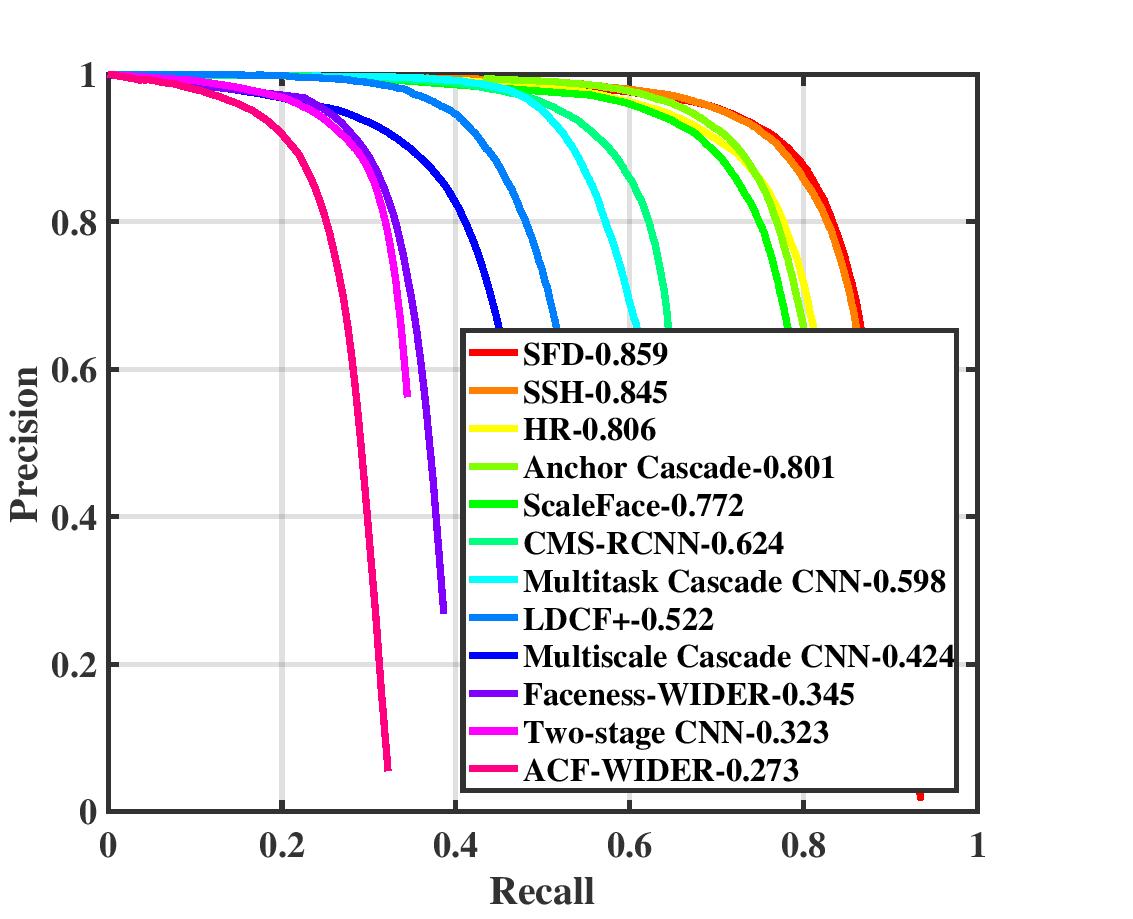}%
\label{fig:exp:wider:hard}}
\caption{The performance of three-stage anchor cascade on WIDER FACE.}
\label{fig:exp:wider}
\end{figure*}

\begin{table}
\caption{The performance of anchor cascade on WIDER FACE. We use min-face $S_{min}=8$.}
\label{tab:exp:rnet96:wider}
\begin{center}
\begin{tabular}{|l|c|c|c|c|c|}
\hline
Method & $n_C$ & Easy & Medium & Hard \\
\hline
MTCNN \cite{zhang2016joint} & - & 0.848 & 0.825 &0.598\\
\hline
HR-ResNet101 \cite{hu2016finding}& - & 0.925 & 0.910 & 0.806 \\
\hline
\hline
APN24+RNet48 & 3 & 0.883 & 0.879 & 0.761 \\
\hline
APN24+RNet48+RNet96 & 3 & 0.906 & 0.895 & 0.801 \\
\hline
\end{tabular}
\end{center}
\end{table}

\subsection{Three-stage Anchor Cascade}

\begin{figure}
\begin{center}
   \includegraphics[width=0.9\linewidth]{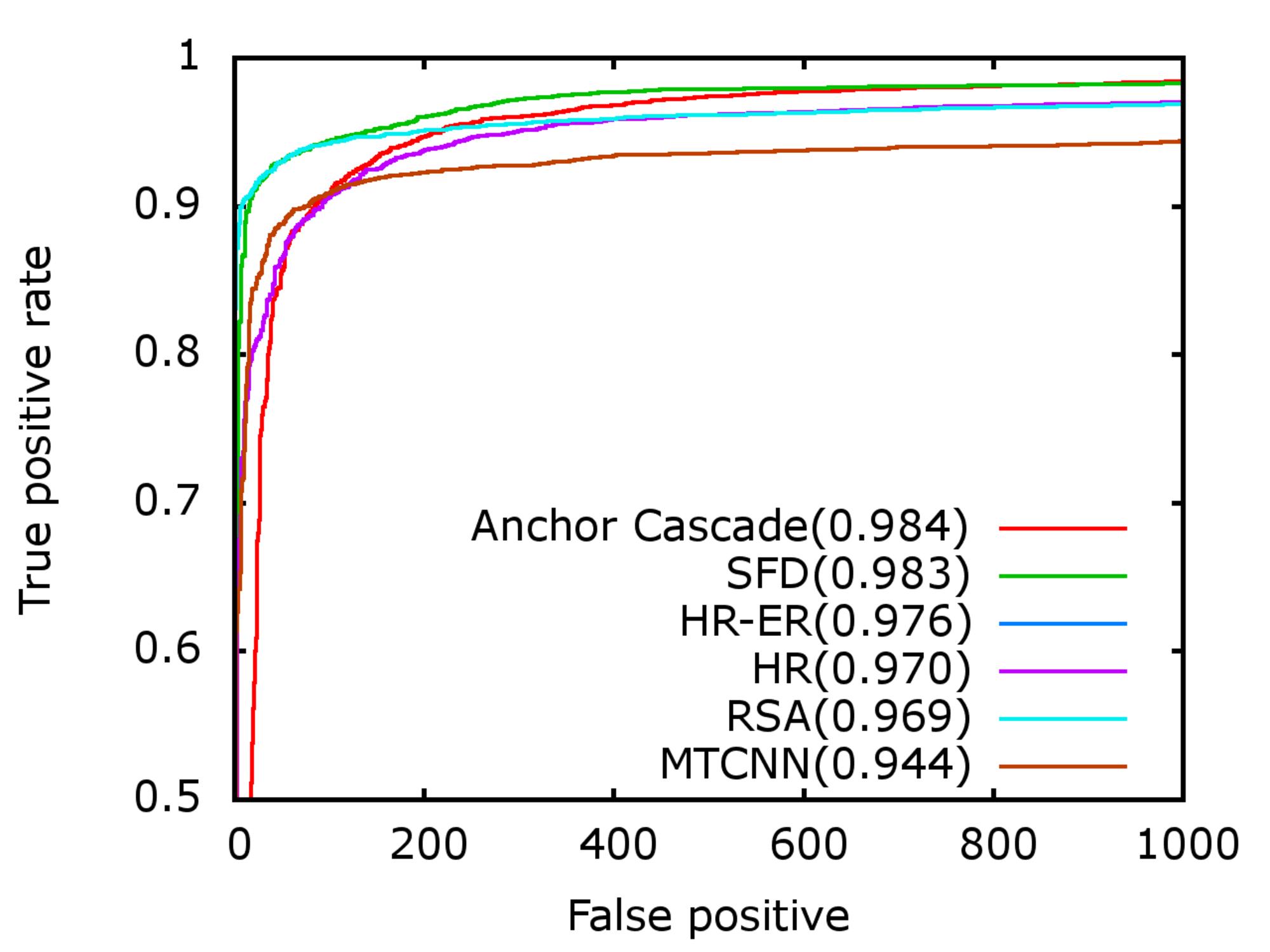}
\end{center}
\caption{The performance of three-stage anchor cascade on FDDB.}
\label{fig:exp:fddb}
\end{figure}

Although the two-stage anchor cascade face detector achieves a good trade-off between detection accuracy and inference speed, we find that most of false positives come from detecting tiny faces. To further boost the performance, we use one more refinement stage RNet96 as the three-stage anchor cascade face detector and demonstrate its performance in Table \ref{tab:exp:rnet96:wider}. We show that the three-stage anchor cascade face detector achieves comparable detection accuracy with HR \cite{hu2016finding} on hard set of WIDER FACE. In Fig. \ref{fig:exp:fddb}, we also find that it achieves a detection rate $98.37\%$ at 1k false positives on FDDB, which outperforms both SSH \cite{najibi2017ssh} ($98.1\%$) and SFD\footnote{They add 238 unlabelled faces as ground truth.} \cite{zhang2017s3fd} ($98.26\%$). However, for detecting tiny faces, there is still a clear margin between anchor cascade  face detectors and the state-of-art anchor-based detectors such as SSH \cite{najibi2017ssh} and SFD \cite{zhang2017s3fd}, as shown in Fig. \ref{fig:exp:wider}. A possible reason is that the anchor cascade face detectors use limited contextual information for efficient face detection, while it is more effective for tiny faces to explore more contextual information, such as body part \cite{zhu2017cms,Zhang_2017_ICCV}.

\section{Conclusion}
In this paper, we propose an anchor cascade framework for efficient face detection by exploring the multi-scale anchors in CNN-based cascade face detectors. To further improve the recall rate, we devise a context pyramid maxout mechanism in harmony with the anchor cascade framework. By using anchor cascade face detector, we further bridge the gap between anchor-based face detectors and CNN-based cascade face detectors. Specifically, our anchor cascade face detector is comparable with typical CNN-based cascade face detectors, e.g., MTCNN, in both model size and running speed, while the detection accuracy has been greatly improved, e.g., from 0.9435 to 0.9704 at 1k false positives on FDDB. Experimental results on FDDB and WIDER FACE demonstrate the effectiveness of the proposed anchor cascade framework for efficient face detection.

\section*{Acknowledgment}

The authors would like to thank...

% Can use something like this to put references on a page
% by themselves when using endfloat and the captionsoff option.
\ifCLASSOPTIONcaptionsoff
  \newpage
\fi

% trigger a \newpage just before the given reference
% number - used to balance the columns on the last page
% adjust value as needed - may need to be readjusted if
% the document is modified later
% \IEEEtriggeratref{8}
% The "triggered" command can be changed if desired:
%\IEEEtriggercmd{\enlargethispage{-5in}}

% references section

% can use a bibliography generated by BibTeX as a .bbl file
% BibTeX documentation can be easily obtained at:
% http://mirror.ctan.org/biblio/bibtex/contrib/doc/
% The IEEEtran BibTeX style support page is at:
% http://www.michaelshell.org/tex/ieeetran/bibtex/
\bibliographystyle{IEEEtran}
% argument is your BibTeX string definitions and bibliography database(s)
\bibliography{ref}
%
% <OR> manually copy in the resultant .bbl file
% set second argument of \begin to the number of references
% (used to reserve space for the reference number labels box)
%\begin{thebibliography}{1}
%
%\bibitem{IEEEhowto:kopka}
%H.~Kopka and P.~W. Daly, \emph{A Guide to \LaTeX}, 3rd~ed.\hskip 1em plus
%  0.5em minus 0.4em\relax Harlow, England: Addison-Wesley, 1999.
%
%\end{thebibliography}

% biography section
% 
% If you have an EPS/PDF photo (graphicx package needed) extra braces are
% needed around the contents of the optional argument to biography to prevent
% the LaTeX parser from getting confused when it sees the complicated
% \includegraphics command within an optional argument. (You could create
% your own custom macro containing the \includegraphics command to make things
% simpler here.)
%\begin{IEEEbiography}[{\includegraphics[width=1in,height=1.25in,clip,keepaspectratio]{mshell}}]{Michael Shell}
% or if you just want to reserve a space for a photo:
\newpage
\begin{IEEEbiography}{Baosheng Yu}
Biography text here.
\end{IEEEbiography}

% if you will not have a photo at all:
\begin{IEEEbiography}{Dacheng Tao}
Biography text here.
\end{IEEEbiography}

% insert where needed to balance the two columns on the last page with
% biographies
%\newpage

%\begin{IEEEbiographynophoto}{Jane Doe}
%Biography text here.
%\end{IEEEbiographynophoto}

% You can push biographies down or up by placing
% a \vfill before or after them. The appropriate
% use of \vfill depends on what kind of text is
% on the last page and whether or not the columns
% are being equalized.

%\vfill

% Can be used to pull up biographies so that the bottom of the last one
% is flush with the other column.
%\enlargethispage{-5in}

% that's all folks
\end{document}